\documentclass[runningheads]{llncs}
\usepackage{graphics} 
\usepackage{epsfig} 
\usepackage{mathptmx} 
\usepackage{times} 
\usepackage{amsmath} 
\usepackage{amssymb}  
\usepackage{amsmath}
\usepackage{bm}
\usepackage{multicol}
\usepackage[ruled,linesnumbered,algo2e]{algorithm2e}
\usepackage[space,sort]{cite}

\SetKw{Kwin}{$\in$}
\SetKw{Kwpara}{in parallel}
\SetKw{KwOp}{Output:} 
\SetKwRepeat{Do}{do}{while}
\newlength\myindent
\usepackage{wrapfig}
\usepackage[figuresright]{rotating}
\usepackage{amssymb}
\usepackage[bookmarks=false]{hyperref}
    \hypersetup{colorlinks,
      linkcolor=blue,
      citecolor=blue,
      urlcolor=blue}

\usepackage{nomencl}
\usepackage{multirow}
\usepackage{graphicx}
\usepackage{textcomp}
\usepackage{lipsum}
\usepackage{xcolor}
\usepackage{array, booktabs, colortbl, tabularx}
\usepackage{stfloats}
\usepackage{afterpage}

\SetAlgorithmName{Algorithm}{Algorithm}

\usepackage{algpseudocode}

\algnewcommand\aand{\textbf{and}}
\algnewcommand\oor{\textbf{or}}

\usepackage{mathtools}

\begin{document}
\title{Real-World Airline Crew Pairing Optimization: Customized Genetic Algorithm versus Column Generation Method}
\titlerunning{Customized Genetic Algorithms for Small \& Complex Flight Network}
%
\author{Divyam Aggarwal\inst{1}
\and
Dhish Kumar Saxena\inst{2}
\and
Thomas B\"ack\inst{3}
\and Michael Emmerich\inst{3}
}
\authorrunning{D. Aggarwal et al.}
%
\institute{Optym, Whitefield, Bengaluru, Karnataka-560048, India \email{divyam.aggarwal@optym.com} \and
Indian Institute of Technology (IIT) Roorkee, Uttarakhand-247667, India \email{dhish.saxena@me.iitr.ac.in} \and Leiden University, 2333 CA Leiden, Netherlands\\ \email{\{t.h.w.baeck, m.t.m.emmerich\}@liacs.leidenuniv.nl}}
\maketitle              
\begin{abstract}

\textit{Airline crew pairing optimization problem} (CPOP) aims to find a set of flight sequences (\textit{crew pairings}) that cover all flights in an airline’s highly constrained flight schedule at \textit{minimum} cost. Since crew cost is second only to the fuel cost, CPOP solutioning is critically important for an airline. However, CPOP is NP-hard, and tackling it is quite challenging. The literature suggests, that when the CPOP's scale and complexity is reasonably limited, and an enumeration of all crew pairings is possible, then Metaheuristics are used, predominantly \textit{Genetic Algorithms} (GAs). Else, \textit{Column Generation} (CG) based Mixed Integer Programming techniques are used. Notably, as per the literature, a maximum of 45,000 crew pairings have been tackled by GAs. In a significant departure, this paper considers over 800 flights of a US-based large airline (with a monthly network of over 33,000 flights), and tests the efficacy of GAs by enumerating all 400,000+ crew pairings, apriori. Towards it, this paper proposes a domain-knowledge-driven customized-GA. The utility of incorporating domain-knowledge in GA operations, particularly \textit{initialization} and \textit{crossover}, is highlighted through suitable experiments. Finally, the proposed GA's performance is compared with a CG-based approach (developed in-house by the authors). Though the latter is found to perform better in terms of solution's cost-quality and run time, it is hoped that this paper will help in better understanding the strengths and limitations of domain-knowledge-driven customizations in GAs, for solving combinatorial optimization problems, including CPOPs.

%
\end{abstract}
\section{INTRODUCTION} \label{intro}
In Airline Scheduling Process, Airline Crew Scheduling (CS) is considered as one of the most important planning activities, since crew operating cost is the second largest after the fuel cost and even its marginal improvements may translate to millions of dollars annually. Over the past three decades, the Operations Research (OR) Society has given unprecedented attention to airline CS and proposed numerous optimization-based solution approaches. To meet the exponentially increasing demand over these years, the expansion of airline operations has lead to a tremendous increase in the number of flights, aircraft, and crew members to be scheduled, leaving the state-of-the-practices obsolete. Given this, it has become imperative to improve existing practices by leveraging recent technological advancements and enhanced computational resources.
\par Airline crew scheduling is a combination of challenging (\textit{NP-hard}~\cite{1}) combinatorial optimization problems, namely, \textit{crew pairing optimization} and \textit{crew assignment} problems, which are tackled sequentially. The former problem aims to generate a set of flight sequences (each called a \textit{crew pairing}) to cover all given flights at \textit{minimum} cost, while satisfying several \textit{legality} constraints linked to the federations' rules, airline-specific regulations, labor laws, etc. The latter problem aims to assign crew members to these optimally-generated pairings while satisfying the pairing and crew requirements. The scope of this research is limited to Airline Crew Pairing Optimization Problem (CPOP). Interested readers are referred to Aggarwal~et~al.~\cite{aggarwal2017interdependence} for a comprehensive review of the integration of other components of the airline scheduling process.
\par In CPOP, crew pairings have to satisfy multiple constraints to be classified as \textit{legal}, and it is imperative to generate legal pairings in a time-efficient manner to assist the subsequent optimization search. Several legal pairing generation approaches, either based on a flight-network or a duty-network, have been proposed in the literature~\cite{14}. Based upon the scale of the CPOP being tackled, the pairing generation module can be invoked using two possible architectures-- one wherein all pairings are enumerated a priori CPOP-solutioning, and the other wherein pairings are enumerated as and when required during the CPOP-solutioning. Regarding solution-methodologies, \textit{mathematical programming techniques} and \textit{metaheuristics}, are commonly employed. In the former category, \textit{Column Generation} (CG)~\cite{lubbecke2010column,lubbecke2005selected} is the most widely adopted technique, which is proven for efficiently solving large-scale CPOPs. It is an efficient search-space exploration technique, that iteratively generates only the pairings having a high potential of bringing in the associated cost benefits. In that, the original CPOP is relaxed into a Linear Programming Problem (LP/LPP); which is then solved iteratively by invoking an LP solver and generating new pairings by solving the corresponding pricing sub-problem(s)~\cite{irnich2005shortest,lubbecke2010column}. Finally, the resulting LPP solution is integerized using an integer programming (IP/IPP) solver or connection-fixing heuristics~\cite{parmentier2020aircraft,5}. For more details, interested readers are referred to~\cite{aggarwal2020airline,desaulniers2020dynamic,4,5}.
\par Among meta-heuristics, the most successful and widely adopted technique is \textit{Genetic Algorithms} (GAs), which are population-based probabilistic-search heuristics, inspired by the theory of natural evolution~\cite{6}. GAs with customized operators are known to be successful in solving a variety of combinatorial optimization problems~\cite{7,18,maskooki2022customized,mittal2020innovative}. Several GA-based CPOP solution approaches, proposed in the literature, are broadly reviewed in Table~\ref{overview}.~\cite{4} is the first instance to customize a GA (using guided GA-operators) for solving a general class of SCPs. In that, the authors validated their proposed approach on small-scale synthetic test cases (with over 1,000 rows 
\begin{table}[htbp]
\caption{An overview of the GA-based CPOP solution approaches from the literature}
\footnotesize
\begin{center}
\begin{tabular}{cccccc}
\hline
 \multirow{1.25}{*}{\textbf{Literature}} & \multirow{2}{*}{\textbf{Formulation$^+$}} & \multirow{1.25}{*}{\textbf{Airline}} & \multicolumn{2}{c}{\textbf{Flight Data*}} & \multirow{2}{*}{\textbf{Airlines}} \\
 \cline{4-5}
 \textbf{Instances} & & \textbf{Timetable} & \textbf{\# Flights} & \textbf{\# Pairings} & \\
 \hline
~\cite{7} & SCP & Did not solve CPOP & 1,000 & 10,000 & - \\
~\cite{8} & SPP & - & 823 & 43,749 & - \\
~\cite{9} & SCP & Daily & 380 & 21,308 & Multiple Airlines \\
~\cite{10} & SCP & Monthly & 2,100 & 11,981 & Olympic Airways \\
~\cite{11} & SCP & Monthly & 710 & 3,308 & Turkish Airlines \\
~\cite{12} & - & - & 506 & 11,116 & Turkish Airlines \\
~\cite{13} & SCP & - & 714 & 43,091 & Turkish Airlines \\
\hline
\end{tabular}
\scriptsize{\\$^+$ SCP stands for Set-Covering Problem formulation and SPP stands for Set-Partitioning Problem formulation. * The provided values are the maximum among all the test-cases being used for validation. \vspace{-5mm}}
\end{center}
\label{overview}
\end{table}
and just 10,000 columns). Notably, the literature review in the table could be summarised in two-fold. First, GAs presented by some instances--~\cite{7,8,9,11,12}, have been validated using the flight networks of smaller airlines, operating in low-demand regions such as Greece, Turkey, etc. (leading to only a handful of all possible legal pairings, up to 45,000 pairings). These GAs become obsolete when scaled to even small flight networks of bigger airlines, operating in large geographical regions such as the USA, etc. Second, the results presented in some of these instances--~\cite{10,13}, have been obtained by solving CPOPs formulated using only a subset of the original search-space (up to 12,000 pairings), i.e., all possible legal pairings are not used. In addition,~\cite{13} demonstrated that despite customizations, GAs failed to solve large-scale CPOPs with the same search-efficiency as small-scale CPOPs. Hence, it is imperative to develop GAs that can efficiently tackle CPOPs with bigger pairing-space, say up to a million.
\par In a significant departure from the existing GA-based approaches, this paper proposes a domain-knowledge-driven customized GA to efficiently tackle a CPOP with over 800 flights of a US-based large airline (operating over 33,000 monthly flights), by enumerating all possible crew pairings (over 400,000 pairings) a priori. In that, the GA operations, particularly initialization and crossover, are enhanced using domain-knowledge. Through suitable experiments, it is demonstrated that the proposed-GA is able to generate crew pairing solutions with varying characteristics such as low number of deadhead flights, crew-hotel-nights, etc., which are important KPIs used by airlines along with the crew pairing cost to evaluate the performance of their schedules. Another contribution of this paper is the insights shared on how well the proposed GA performs in comparison to a mathematical programming-based CPOP solution approach, on which the literature is mostly silent upon. For this comparison, a CG-based large-scale airline crew pairing optimizer (\textit{CG-Optimizer}), developed in-house by the authors and validated by the research consortium's industrial sponsor-- GE Aviation has been utilized. Though the CG-Optimizer is found to perform well in terms of the solution's cost quality and runtime, it is hoped that this paper will help better understand the strengths and limitations of domain-knowledge-driven customizations in GAs, for solving challenging combinatorial problems like CPOPs.
\section{Airline Crew Pairing Optimization Problem}
\begin{figure}[htbp]
\vspace{-3mm}
  \centering
  \framebox{\parbox{0.7\columnwidth}{\includegraphics[width=0.7\columnwidth, keepaspectratio]{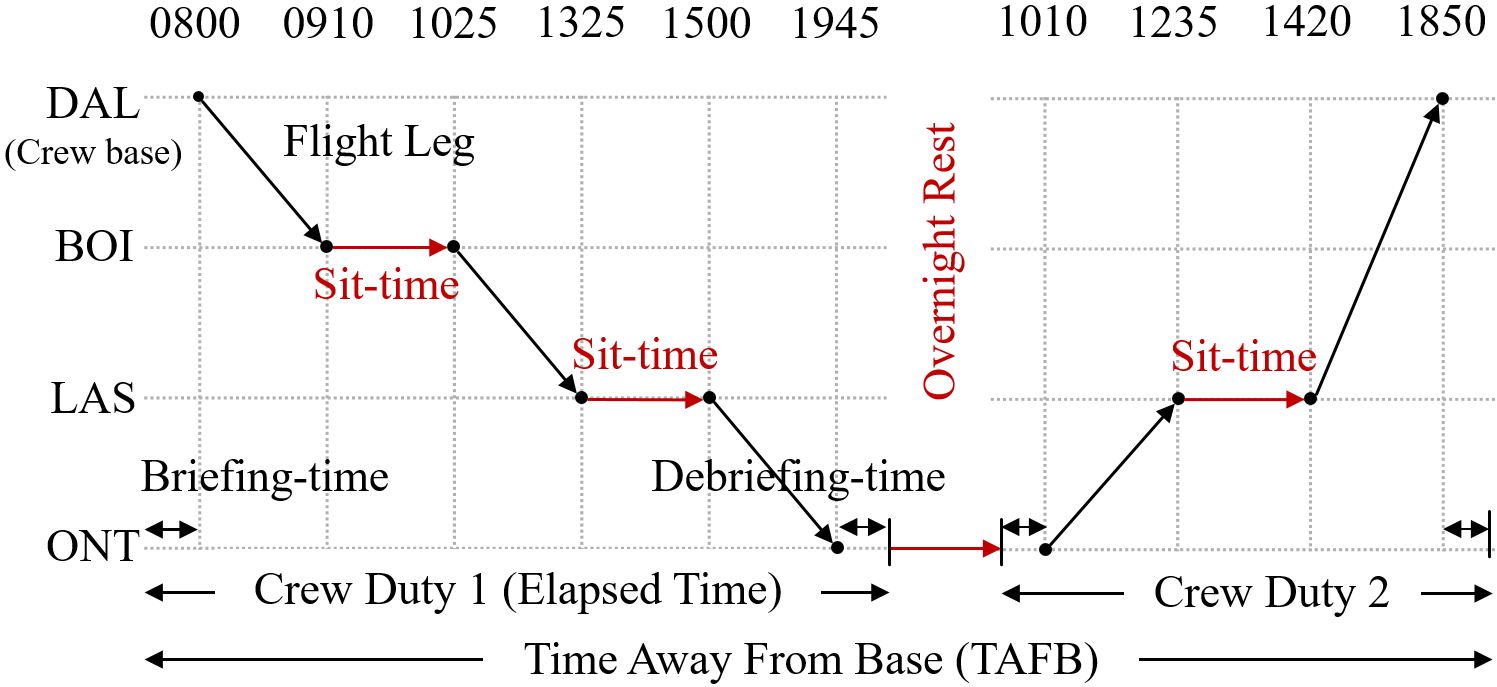}}}
  \caption{A crew pairing beginning from \textit{Dallas} (\textit{DAL}) crew base}
  \label{crewpair}
\end{figure}

\vspace{-1mm}
In CPOP, the input data includes an airline flight schedule with a finite number of flights, the pairings' costing criterion, and legality rules \& regulations. As introduced before, a \textit{crew pairing} is a sequence of flights to be flown by a crew member, beginning and ending at the same crew base. Other associated terminologies of CPOP are explained with the help of an example of a crew pairing, shown in Fig.~\ref{crewpair}. At times, a crew is required to be transported to an airport to fly their next flight. In such situations, the crew is transported as passengers in another flight, flown by another crew. Such a flight is called a \textit{deadhead} or a \textit{deadhead flight} for the transported crew. The presence of deadhead flights affects an airline’s profit in two ways. First, the airline has to bear the loss of revenue on the passenger seats being occupied by the deadhead-ing crew, and the other is it has to pay the hourly wages to the deadhead-ing crew even when they are not servicing the flight. To maximize profits, airlines desire to minimize these deadheads as much as possible (ideally zero).
\par As mentioned in Section~\ref{intro}, it is imperative to develop a \textit{legal crew pairing generation} approach to facilitate legal pairings to the optimization phase. In small- and medium-scale CPOPs, all legal pairings are generated explicitly before the optimization phase. The same approach is adopted in this work, and a duty-network-based parallel legal pairing generation algorithm~\cite{14} is used for generating all legal pairings explicitly. Interested readers are referred to~\cite{14} for an extensive review of the pairing generation literature too.
\par The goal of the optimization phase is to find a pairing subset from the generated set of all legal pairings to cover the given flights with the minimum cost possible. In literature, the CPOP is modeled either as a set-partitioning problem (SPP; each flight leg is allowed to be covered only once) or as a set-covering problem (SCP; over-coverage of flight legs i.e. deadheads are allowed). In this paper, the SCP formulation is adapted and modified to define the optimization problem for the proposed GA. Its mathematical model is presented in Section~\ref{fitness}.
\section{Genetic Algorithm} \label{sec:GA}
A customized-GA is proposed in this work to solve CPOPs for which enumeration and handling of the entire pairing set is computationally-tractable. Before starting the GA-search, the entire pairing set, denoted by $AllPairings$, is enumerated \textit{a priori}. After pairing enumeration, the proposed GA tackles CPOP by formulating a fitness function based on its SCP formulation (given in Section~\ref{fitness}). The GA-search starts by initializing the population using efficient pairing sets (each set $\subset AllPairings$) by employing a novel initialization heuristic. Iteratively, the population is improved upon, each iteration referred to as a \textit{generation}, by bringing-in new pairings from the remaining pairings' space using enhanced genetic operators.
\par The working of the proposed GA is explained in conjunction with the enhancement of its genetic operators, as described in the upcoming subsections. Notably, these genetic operators have either been enhanced or adopted from the GA-variants proposed in~\cite{7,10,15}. Now, the high-level pseudocode of the proposed GA, formalized in lines 1-13 of Algorithm~\ref{PseudoCode}, is explained below. In line 2, a set of chromosomes, notated as $InitialPop$, is generated by applying the novel initialization heuristic on $AllPairings$. In line 3, the fitness function value of chromosomes $\in InitialPop$ is computed. Lines 4-12 constitute GA-generations, which terminate as soon as the user-specific termination criterion is satisfied. This is followed by the selection of best-fit chromosomes $\in InitialPop$ (for generation = 1) or $\in BestPop$ (for generations $>$ 1) that constitutes the parent population, notated as $ParentPop$ (line 5). Subsequently, in line 6, the parent chromosomes reproduce to generate child chromosomes (set notated as $ChildPop$) via crossover operation. In line 7, child chromosomes $\in ChildPop$ are mutated to promote diversity in the solutions' pool. Notably, here, two different crossover and mutation operators are interchangeably used according to different settings of the proposed GA, as mentioned in Section~\ref{sec:CompExps}. Being a combinatorial optimization problem, the generated child chromosomes may be infeasible with respect to flight coverage constraints. To re-install their feasibility, a \textit{feasibility-repair heuristic} along with a \textit{redundant-pairing removal heuristic} is applied to $ChildPop$ (lines 8-9). Next, the fitness of chromosomes $\in ChildPop$ is computed (line 10). Finally, the chromosomes $\in ChildPop$ are combined with chromosomes $\in ParentPop$ using a population replacement operator, resulting in $BestPop$ and forming the input for the next generation.
\subsection{Novel Chromosome Representation} \label{sec:chromosome}
\begin{figure}[htbp]
	\centering
	\includegraphics[width=0.7\columnwidth, keepaspectratio]{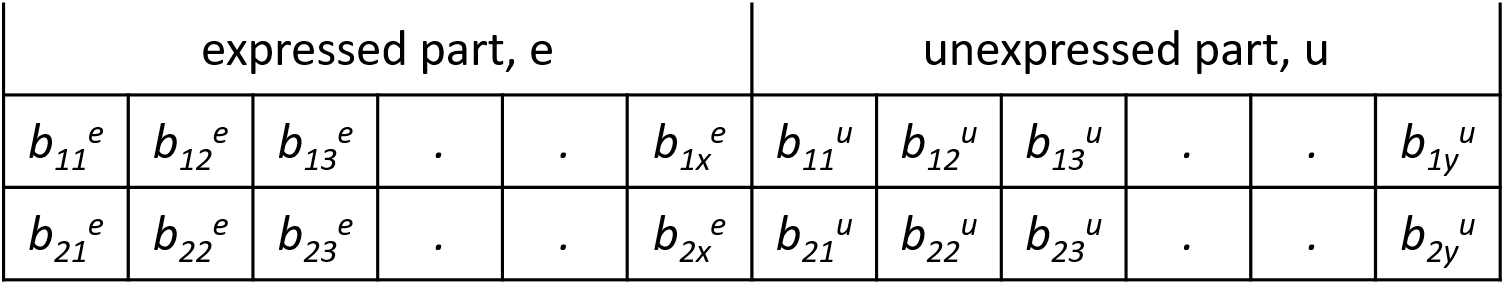}
	\caption{Chromosome structure}
	\label{chromosome}
	\vspace{-5mm}
\end{figure}

The proposed GA utilizes an architecture wherein all possible legal pairings, set denoted by $AllPairings$, are enumerated explicitly. Given that $AllPairings$ may contain thousands of pairings, conventional binary-chromosome structure, containing genes corresponding to each pairing, will become impractical. As a result, a chromosome with 2-bits gene-encoding is proposed here, whose structure is illustrated in Figure~\ref{chromosome}. In that, the first bit, notated as $b_{1i}$, is an integer representing the index of a pairing $p_i \in AllPairings$, which constitutes the chromosome. And the second bit, notated as $b_{2i}$, is a binary number representing the participation of the corresponding pairing for fitness evaluation ($=1$, or $=0$ otherwise). Moreover, to maintain diversity in the chromosome and prevent premature convergence, the chromosome structure, used in~\cite{15}, is adapted here.
\begin{algorithm2e}
\DontPrintSemicolon
\footnotesize{
\Begin{
\textbf{Proposed GA:}\;
$InitialPop \gets$ Call Minimal-deadhead Initialization Heuristic($AllPairings$)\;
Evaluate Fitness of $InitialPop$\;
\While{Termination criterion is not met}{
$ParentPop \gets$ Selection operator($InitialPop/BestPop$)\;
\ \ $(Child1,\ Child2) \gets$ Crossover($Parent1,\ Parent2$)\ \ \ \ \ \ \ \ \ \ \ \ \ \ \ \ \ \ \ \ \ \ \ \ \ \ \ \ \ \ \ \ \ \ \ \ \ \ \ \ \ \ \ \ \ \ \ \ \ \ \ \ \ \ \ \ \ \ \ \ \ \ \ \ \ \ \ \ \ \ \ \ \ \ \ \tcc{\textit{Crossover1 or Crossover2}}
Mutation($ChildPop$)\ \ \ \ \ \ \ \ \ \ \ \ \ \ \ \ \ \ \ \ \ \ \ \ \ \ \ \ \ \ \tcc{\textit{Mutation1 or Mutation2}}
Feasibility-repair Heuristic($ChildPop$)\;
\ \ \ \ \ \ Redundant-pairing Removal Heuristic($ChildPop$)\;
Fitness Evaluation($ChildPop$)\;
$BestPop \gets$ Population Replacement($ParentPop \cup ChildPop$)\;
}
\textbf{Minimal-deadhead Initialization Heuristic:}\;
\ForEach{chromosome $\in InitialPop$}{
\For{expressed part}{
Randomly select a zero-deadhead solution from $AllPairings$\;
\uIf{|all flights \textbf{are not} covered|}{
Select pairings from $AllPairings$ w.r.t. the number of deadheads they are bringing into the solution\;
}}
\For{unexpressed part}{
Randomly select pairings from $AllPairings$ without replacement
}}
\textbf{Crossover2:}\;
$CombinedPairings \gets$ Combined list of pairings in $Parent1$ and $Parent2$\;
\ForEach{\textit{child} chromosome}{
\For{expressed part}{
Randomly select a zero-deadhead solution from $CombinedPairings$
}
\For{unexpressed part}{
Select pairings from $CombinedPairings$ based on their dissimilarity with expressed part (number of non-intersecting flights)}
}
\textbf{Procedure} Redundant-pairing Removal Heuristic ($ChildPop$):\;
\ForEach{pairing corresponding to index $b_{1i}^e$ of each chromosome $\in ChildPop$}{
\uIf{|$b_{2i}^e$ = 1|}{
set $b_{2i}^e$ = 0\;
\uIf{|all flights are \textbf{not covered} in chromosome}{
set $b_{2i}^e$ = 1
}}}
}
}
\caption{Pseudocode of the proposed GA and its constituting operators}
\label{PseudoCode}
\end{algorithm2e}
As a result, a chromosome contains two parts, namely \textit{expressed} and \textit{unexpressed} parts, notated as $B^e: (b^e_1, b^e_2)$ and $B^u: (b^u_1, b^u_2)$, respectively. The former part involves pairings that participate in the fitness evaluation of the chromosome, whereas the latter part involves pairings that are not considered. However, these pairings ($\in$ unexpressed part) are utilized to preserve diversity with respect to the pairings in expressed part, so that diverse pairings can participate in the reproduction of child chromosomes. Moreover, contrary to the chromosome structure used in~\cite{15}, the length of the chromosome is kept fixed, and the length of expressed and unexpressed parts is allowed to vary during generations, making this a novel adaptation.
\subsection{Fitness Evaluation} \label{fitness}
CPOP aims to minimize total crew pairing cost while covering all flights by at least one pairing. To evaluate the fitness of a chromosome, the fitness function is constructed using the set-covering problem formulation (SCP) of the airline CPOP~\cite{aggarwal2020airline,aggarwal2020novel}, which is given as follows:
\begin{equation}
\label{obj}
\displaystyle min \left\{\sum_{b^e_{1j} \in b_1^e} c_j . b^e_{2j} + \psi_D . \sum_{i \in S_F} \left(\sum_{b^e_{1j} \in b_1^e} a_{ij} . b^e_{2j} - 1\right)\right\}
,\ s. t. \sum_{b^e_{1j} \in b_1^e} a_{ij} . b^e_{2j} \geq 1 \ \forall i \in S_{F}
\end{equation}
In that, Eq.~\ref{obj} represents the objective function and feasibility constraint of the airline CPOP. In that, $S_F$ is the flight set to be covered; $c_j$ is the cost of pairing $p_j$ given by $b^e_{1j}$; $\psi_D$ is the deadhead penalty-cost set by airlines; $a_{ij}$ is a binary constant representing the coverage of flight $f_i$ by pairing $p_j$ ($=1$, or $=0$ otherwise); and $b_{2j}^e$ is the binary decision variable (given by the binary bit of the expressed part of the chromosome), representing the selection of pairing $p_j$ in the solution ($=1$, or $=0$ otherwise). Notably, the pairing and deadhead costs (objective function components in Eq.~\ref{obj}) are two objectives to be minimized. Here, a weighted scalarization approach is used for combining them, wherein deadhead flights are penalized using the penalty cost assigned by the industrial sponsors.
\subsection{Minimal-deadhead Initialization Heuristic} \label{sec:InitMethod}
Generally, chromosomes in the initial population are generated using randomly selected genes to allow for exploratory search upfront. However, in single-objective problems like CPOP, it is imperative to constitute the initial population with diverse and reasonably good-quality chromosomes, supporting the initial-exploration stage while expediting the convergence. In this work, an effective initialization heuristic, referred to as \textit{Minimal-deadhead Initialization Heuristic} (lines 14-24 of Algorithm~\ref{PseudoCode}), is proposed, which induces pairings that bring a lesser number of deadhead-flights into the solution. To generate the expressed part, first, a zero-deadhead pairing set ($\subset AllPairings$) is chosen randomly (lines 16-20). If all flights are not covered in the expressed part, then more pairings (from $AllPairings$) are introduced in the increasing order of the number of deadhead flights they bring in. The unexpressed part is initialized using randomly selected pairings from $AllPairings$ until the finite length of the chromosome is exhausted (lines 21-23). \vspace{-1mm}
\subsection{Selection} \label{sec:selection}
This operator selects parent chromosomes from the input population according to their fitness-function values. Here, a binary tournament selection operator~\cite{goldberg1991comparative} is utilized. It creates $N$ sets of two randomly-selected chromosomes ($N$ being the population size) and selects the fittest chromosome to constitute the resulting parent population-- $ParentPop$. \vspace{-1mm}
\subsection{Crossover} \label{sec:cross}
In crossover, new child chromosomes are reproduced by transforming genetic information from parent chromosomes using different strategies, such as one-point crossover, two-point crossover, uniform crossover, fusion crossover~\cite{7}, etc. Here, two specific crossovers are studied and compared, namely \textit{Crossover1} and \textit{Crossover2}. The former is the fusion crossover~\cite{7} that has been widely adopted in CPOP's literature. In that, a fitness-based probability is used to decide the gene of which parent chromosome will pass on to the child chromosome. The latter is an adaption of a domain-knowledge-driven greedy crossover~\cite{15}, which was originally proposed to improve the convergence of the GA-search. Its pseudocode is given in lines 26-35 of Algorithm~\ref{PseudoCode}. In that, the child chromosome's expressed part is constructed using a zero-deadhead pairing set, which is selected randomly from the combined pool of pairings in the parent chromosomes. And, its unexpressed part is formed using the remaining pairings on the basis of their dissimilarity with the expressed part, i.e., the flights they are covering differently compared to the expressed part. \vspace{-1mm}
\subsection{Mutation} \label{mutate}
In mutation, certain genes of the child chromosomes (from crossover) are altered to prevent premature convergence. Here, two widely-adopted mutation strategies are studied and compared. The first is a bit-flip mutation, referred to as \textit{Mutation1}, and the other is the mutation proposed in~\cite{10}, referred as \textit{Mutation2}, which utilizes density of the fittest solution in the population. In \textit{Mutation1}, if an $i^{th}$ gene gets selected for mutation, then $b_{2i}$ bit is flipped from 0 to 1 or vice-versa. Whereas in \textit{Mutation2}, if an $i^{th}$ gene is selected for mutation, then $b_{2i}$ is mutated from 0 to 1 or vice-versa, based on a probability equivalent to the percentage of 1s in the fittest individual. \vspace{-1mm}
\subsection{Feasibility-Repair Heuristic}
It is well-known that in combinatorial optimization problems such as CPOP, crossover and mutation operations may render the child chromosomes infeasible, leading to the requirement of a feasibility-repair step. A heuristic, proposed in~\cite{7}, is adapted in this work by adding a redundant-pairing removal step. In that, for each uncovered flight in the infeasible chromosome, a pairing ($\in AllPairings$) with minimum value of a quality index (defined as \textit{Cost of pairing$/$Number of uncovered flights the pairing covers}) is selected. After this, a redundant-pairing removal heuristic is proposed (lines 37-43 of Algorithm~\ref{PseudoCode}), which finds and removes the pairings with zero contribution in the overall flight coverage of the chromosome.  \vspace{-1mm}
\subsection{Population Replacement}
The last step is the population replacement step wherein $ParentPop$ \& $ChildPop$ are combined to select the $ParentPop$ for next generation, notated as $BestPop$. There exists two main strategies, namely generational and steady-state. Here, generational strategy is adopted in which selects best \textit{N} chromosomes out of \textit{N} parent and \textit{N} child chromosomes to constitute $BestPop$. \vspace{-1mm}
\section{Computational Experiments} \label{sec:CompExps}
All the computational experiments in this research work are performed with a real-world test-case, which includes 839 flights and crew based on a single home base-- Dallas, USA (\textit{DAL}). This test-case has been extracted from the networks of US-based big airlines (operating upto 33,000 monthly flights with upto 15 crew bases), provided by the research consortium's industrial sponsors-- GE Aviation. It is found that 430,873 legal crew pairings are possible for this test-case, which is enormously huge in comparison to the amount of pairings dealt in the existing GA-based approaches (Section~\ref{intro}). In this research work, all the algorithms are implemented using \textit{Python} and executed using just-in time (JIT) compiler-- \textit{PyPy}, improving the computational speeds by a great extent. All computations are performed on a HP Z640 workstation (2 X Intel$^\circledR$ Xeon$^\circledR$ Processor E5-2630v3 @2.40GHz and 8-Cores/16-Threads, enabled with parallelization capabilities). 
\par The parameter settings of the proposed GA, used for the experiments in this research, are given in Table~\ref{paraSet}.
\begin{table}[b]
\vspace{-2mm}
\begin{minipage}{0.5\linewidth}
\centering \footnotesize
\caption{GA parameter settings}
\begin{tabular}{ll}
\hline
\textbf{Parameters} & \textbf{Value} \\
\hline
Population Size & $24$\\
Termination & $5000\ seconds$ \\
Chromosome Length & $100 + MaxLen(InitialPop)$\\
Crossover Rate & $0.9$\\
Mutation Rate & $3 \cdot (1/ChromosomeLen)$\\
\hline
\end{tabular}
\label{paraSet}
\end{minipage}
\hfill
\begin{minipage}{0.45\linewidth}
\centering \footnotesize
\caption{GA configurations}
\begin{tabular}{lcccc}
\hline
 \textbf{Operators} & \textbf{GA1} & \textbf{GA2} & \textbf{GA3} & \textbf{GA4} \\
\hline
Initialization &  & \cellcolor{black!50} & \cellcolor{black!50} & \cellcolor{black!50} \\
Mutation1 & \cellcolor{black!50} & \cellcolor{black!50} & & \\
Mutation2 & & & \cellcolor{black!50} & \cellcolor{black!50} \\
Crossover1 & \cellcolor{black!50} & \cellcolor{black!50} & \cellcolor{black!50} & \\
Crossover2 & & & & \cellcolor{black!50} \\
\hline
\end{tabular}
\label{GAConfigs}
\end{minipage}
\end{table}
It is observed that on increasing the GA's population size, the number of GA-generations may decrease as it may bring more diversity in the population's solution-quality at each generation. However, each generation's time may increase proportionately. Overall, this may not drastically degrade the final runtime-performance. Hence, the population size here is selected accordingly. For the termination of the proposed GA, its overall runtime is selected as the termination criterion instead of the number of generations, given different strategies being used in different settings of GAs being compared here. The chromosome length has been selected in accordance to the best practice solutions. In~\cite{16}, $(1/ChromosomeLen)$ is proposed as the lower bound for the optimal mutation rate. However, during experiments, it is observed that this lower bound shall be inflated by some factor (here, 3) in order to prevent premature convergence. This is also in alignment with the observations of the authors in~\cite{4} with variable mutation rate.
\par
\begin{figure}[t]
  \vspace{-3mm}
  \centering
  \framebox{\parbox{0.8\columnwidth}{\includegraphics[width=0.8\columnwidth, keepaspectratio]{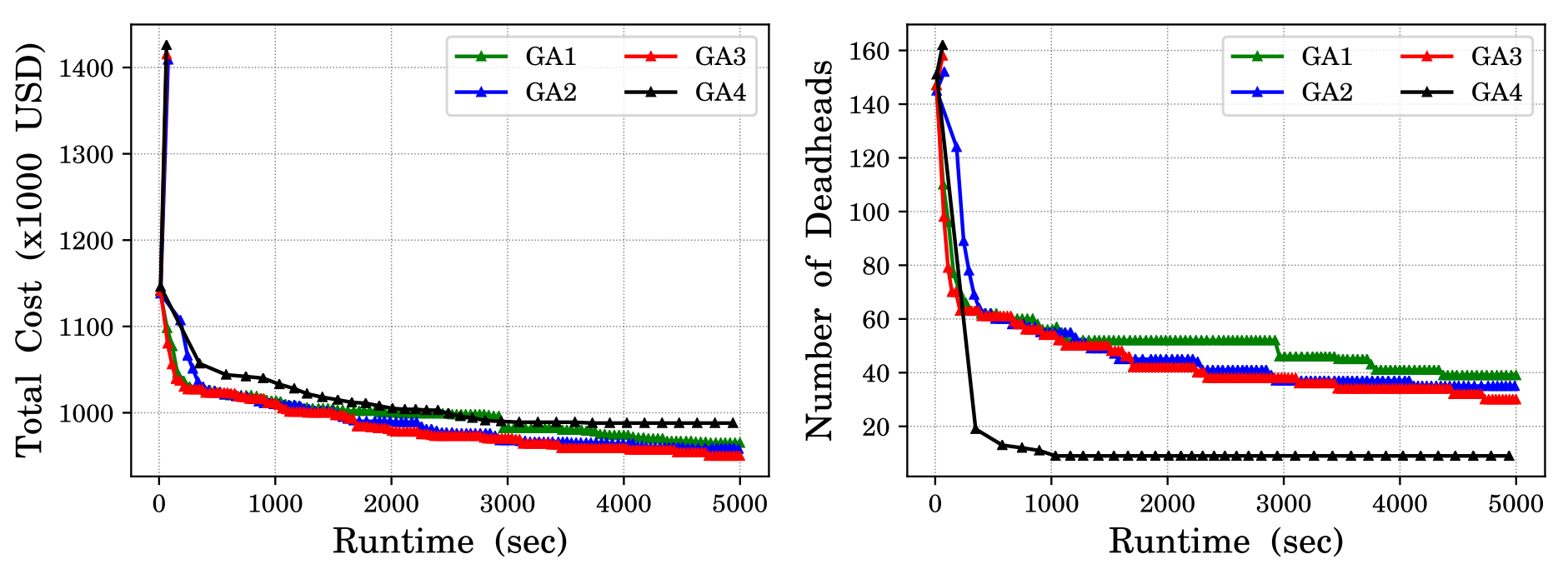}}}
  \caption{Characteristic plots of the GA-runs}
  \label{plot1}
  \vspace{-4mm}
\end{figure}

 In this research work, variants of GA-operators are proposed which are either developed by the authors or adapted from the variants present in the literature. To solve the above-mentioned airline test-case and similar problems, it is imperative to find the most effective combination of these operators. Towards this, four configurations of the GA are implemented and tested in this work, the structure of whom are shown in Table~\ref{GAConfigs}. For each of these GA-configurations, ten runs, initialized with different random seeds (uniformly distributed between 0 and 1), are performed. The experimental results of these runs are summarized in Table~\ref{GAallruns} and the comparative plots are shown in Fig.~\ref{plot1}. First, the merits of using the proposed minimal-deadhead initialization heuristic are assessed. For this, the best solution among the initial populations generated in GA1-runs (using random initialization) and GA2-runs (using the proposed initialized heuristic), are compared, as recorded in first two rows of Table~\ref{GAallruns}. It is observed that the characteristics of the best initial solution from the GA2-runs (number of deadheads and total cost) are reasonably very good compared to those of GA1-runs. Notably, the initialization runtime for these GA-configurations are similar, as the additional runtime consumed by the proposed heuristic is compensated by the runtime required to repair the infeasible solutions obtained using random initialization in GA1-runs. Moreover, GA2-runs lead to a better-cost crew pairing solution (best solution across all seeds) compared to the GA1-runs. These observations endorse the effectiveness of using the proposed initialization heuristic.
\par Second, the merits of the proposed mutation strategies are assessed. For this, the GA-configurations-- GA2 (using \textit{Mutation1}) and GA3 (using \textit{Mutation2}) are compared. From the results recorded in Table~\ref{GAallruns}, it is observed that GA3-runs lead to a better crew pairing solution (in terms of both cost and number of deadheads) compared to the GA2-runs. However, the difference between them is marginal, equalizing the effects of both mutation strategies. Consecutively, \textit{Mutation2} is considered for the subsequent experiments.
\par Third, the merits of the proposed crossover strategies are assessed. For this, the GA-configurations-- GA3 (using \textit{Crossover1}) and GA4 (using \textit{Crossover2}), are compared.
\begin{table}[htbp]
\vspace{-2mm}
\caption{Experimental results of the GA-runs}
\begin{center}
\begin{tabular}{llccccccc}
\hline
\multirow{1.25}{*}{\textbf{Runtime}} & \multirow{2.25}{*}{\textbf{GAs}} & \multicolumn{3}{c}{\textbf{Crew Pairing Cost (USD)}} & & \multicolumn{3}{c}{\textbf{\# Deadheads}} \\
\cline{3-5} \cline{7-9}
 \multirow{0.75}{*}{\textbf{(sec)}} &  & $\pmb{\overline x \pm \sigma}$ & \textbf{Best} & \textbf{Worst} &  & $\pmb{\overline x \pm \sigma}$ & \textbf{Best} & \textbf{Worst} \\
\hline
\multirow{2}{*}{70} & GA1 & 2,649,823 $\pm$ 57,559 & 2,494,649 & 2,710,084 &  & 1,095 $\pm$ 45 & 977 & 1,151 \\
 & GA2 & 1,417,223 $\pm$ 9,380 & 1,398,427 & 1,430,115 &  & 156 $\pm$ 06 & 149 & 164 \\
 \hline
\multirow{4}{*}{5000} & GA1 & 980,226 $\pm$ 23,091 & 964,857 & 1,037,504 & & 40 $\pm$ 04 & 35 & 49 \\
 & GA2 & 1,195,229 $\pm$ 225,555 & 957,832 & 1,430,115 & & 98 $\pm$ 61 & 35 & 164 \\
 & GA3 & 1,192,104 $\pm$ 228,745 & 949,591 & 1,430,115 & & 98 $\pm$ 61 & 30 & 164 \\
 & GA4 & 993,209 $\pm$ 5,337 & 987,638 & 1,001,487 & & 09 $\pm$ 04 & 06 & 21 \\
\hline
\label{GAallruns}
\end{tabular}
\vspace{-10mm}
\end{center}
\end{table}
From the tabulated results and plot of GA4 in Fig.~\ref{plot1}, it is quite evident that the proposed \textit{Crossover2} strategy is highly effective in reducing the number of deadheads that too in a very less runtime.
However, the cost of the final crew pairing solution from GA4-runs is marginally poorer than those of the GA3-runs. On further analyzing the crew pairings of best solution from GA4-runs, it is observed that the majority of pairings contain very less number of flight legs, each referred to as a \textit{short-pairing}. Hence, with such short-pairings, the solution contains a large number of pairings to cover all 839 flights. Moreover, during the GA search, the current best solution, dominated by large number of short-pairings, becomes too rigid to allow any large-pairing (covering a large number of flights, contrary to a short-pairing) to enter the solution, hence, stopping the search at local optima.
\begin{table}[b]
\caption{Crew pairing solutions of CG-Optimizer and proposed GA configurations}
\begin{center}
\begin{tabular}{ccccc}
\hline
\multirow{2}{*}{\textbf{Algorithms}} & \textbf{Total Cost} & \multirow{2}{*}{\textbf{\# Deadheads}} & \multirow{2}{*}{\textbf{\# Pairings}} & \textbf{\%age Cost} \\
 & \textbf{(USD)} &  &  & \textbf{Gap} \\
\hline
\textit{CG-Optimizer} & 850,303 & 02 & 142 & 0\\
GA1 & 964,858 & 39 & 169 & 13.47\\
GA2 & 957,833 & 35 & 172 & 12.65\\
GA3 & 949,592 & 30 & 171 & 11.68\\
GA4 & 987,639 & 09 & 242 & 16.15\\
\hline
\label{bestSols}
\end{tabular}
\vspace{-5mm}
\end{center}
\end{table}
\par As mentioned before, a large-scale column generation based airline crew pairing optimizer (\textit{CG-Optimizer}) is used in this research to assess the performance of the proposed GA-configurations, and to share the insights on how well a highly-customized GA performs in comparison to advanced mathematical programming techniques. \textit{CG-Optimizer} is developed in-house by the authors as part of the overall research project, and has been tested and validated on real-world, large-scale and complex flight networks provided by GE Aviation. The exhaustive details of \textit{CG-Optimizer} are presented in the technical report--~\cite{aggarwal2020airline}. The final crew pairing solution of \textit{CG-Optimizer} is compared with the best solutions of the proposed GA configurations, and the results are recorded in Table~\ref{bestSols}. From the tabulated results, it is quite evident that the crew pairing solution offered by \textit{CG-Optimizer} is of superior quality than any of the solutions offered by the proposed GA configurations, with minimum percentage cost difference being 11.7\%. Moreover, the number of deadheads as well as the number of pairings involved are minimal in the solution offered by \textit{CG-Optimizer}. This endorses the fact that CG-based (mathematical programming) CPOP solution approaches are highly effective in solving CPOPs with moderately sized flight networks, compared to a GA-based solution approach despite several domain-knowledge-driven enhancements. \vspace{-1mm}
\section{Conclusion}
This paper proposes a domain-knowledge-driven customized GA, with enhanced genetic operations, particularly initialization and crossover, to efficiently tackle a CPOP with over 800 flights of a US-based large airline (operating over 33,000 monthly flights), by enumerating all 400,000$+$ crew pairings a priori. The proposed minimal-deadhead initialization heuristic is effective in achieving a better-initial solution compared to a random initialization strategy (with $78\%$ better cost and $555\%$ lesser deadheads) in approximately similar runtime. On assessing the performance of two widely-adopted mutation operators, it is found that both perform similarly with \textit{Mutation2} performing marginally better than \textit{Mutation1}. A deadhead-minimizing crossover operator, \textit{Crossov-er2}, is also proposed which is found to be effective in reducing the number of deadheads significantly within a short runtime.
\par In addition to the above, this paper shares insights on the comparison of customized GAs with CG-based CPOP solution approaches. For this, the performance of the proposed GA is compared viz-a-viz CG-based large-scale optimizer (\textit{CG-Optimizer}, developed by the authors) to solve large-scale CPOPs with over billion-plus legal pairings, 4,000 flights, and 15 crew bases. Though it is found that the crew pairing solution offered by \textit{CG-Optimizer} is of superior quality than any of the solutions offered by the proposed GA (with minimum percentage cost difference being 11.7\%), it is believed that this paper will serve as a template to better understand the strengths and limitations of domain-knowledge-driven customizations in GAs (other metaheuristics) for solving combinatorial optimization problems, including CPOPs.
\par Notably, \textit{Crossover2} favors deadhead minimization, leading to the selection of short-pairings and driving the GA-search towards local optima. Towards it, search-space expansion heuristics~\cite{17}, and variable mutation rates~\cite{7} could be adapted. This work paves the way for a detailed multi-objective study of the airline CPOP under realistic assumptions. Moreover, an important future direction is to investigate the trade-off between crew operating cost and robustness against delays (by adding slack time to the duration of given flights). This slack time can be based on the likelihood of delays (obtained using a machine learning model) and/or based on the systemic importance of flight connections. Lastly, the emergent trend of utilizing machine learning capabilities to assist combinatorial optimization using metaheuristics or mathematical programming may also hold promise to improve the current propositions~\cite{morabit2021machine,aggarwal2020learning,mittal2021learning,morabit2022machine,liu2022good,shen2022enhancing}.
\section{Acknowledgement}
This research work is supported by MEITY, India [grant 13(4)/2015-CC\&BT]; NWO, the Netherlands; and GE Aviation, India. Thanks to the industrial sponsor's (GE Aviation) team members: Saaju Paulose, Arioli Arumugam and Rajesh Alla for their invaluable support in successfully completing this research. Notably, during this research, the first author (Divyam Aggarwal) was a Ph.D. Candidate at IIT Roorkee, India.
%
\bibliographystyle{splncs04}
\bibliography{GAPaperBibtex}
\end{document}